\pdfoutput=1
\documentclass[11pt]{article}

\usepackage{acl}

\usepackage{times}
\usepackage{latexsym}

\usepackage[T1]{fontenc}
\usepackage[utf8]{inputenc}
\usepackage[compact]{titlesec}
\usepackage{lipsum}

\usepackage{microtype}
\usepackage{booktabs}
\title{``It's Not Just Hate'': A Multi-Dimensional Perspective on Detecting Harmful Speech Online}

\usepackage{graphicx}
\author{Federico Bianchi \\
  Stanford University \\
  Stanford, California, USA  \\\And
  Stefanie Anja Hills \\
  University of Stirling \\
   Stirling, UK  \\ \And
  Patricia Rossini \\
  University of Glasgow \\
  Glasgow, UK  \\ \AND 
  Dirk Hovy \\
  Bocconi University \\
  Milan, Italy \\ \And
    Rebekah Tromble \\
  George Washington University  \\
  Washington, DC, USA \\ \And
    Nava Tintarev \\
 Maastricht University \\
 Maastricht, The Netherlands \\
  }

\begin{document}
\maketitle
\begin{abstract}
Well-annotated data is a prerequisite for good Natural Language Processing models. Too often, though, annotation decisions are governed by optimizing time or annotator agreement. We make a case for nuanced efforts in an interdisciplinary setting for annotating offensive online speech. Detecting offensive content is rapidly becoming one of the most important real-world NLP tasks. However, most datasets use a single binary label, e.g., for \textit{hate} or \textit{incivility}, even though each concept is multi-faceted. This modeling choice severely limits nuanced insights, but also performance.
We show that a more fine-grained multi-label approach to predicting incivility and hateful or intolerant content addresses both conceptual and performance issues.
We release a novel dataset of over 40,000 tweets about immigration from the US and UK, annotated with six labels for different aspects of incivility and intolerance.
Our dataset not only allows for a more nuanced understanding of harmful speech online, models trained on it also outperform or match performance on benchmark datasets.\\
\textcolor{red}{\textbf{Warning}:
This paper contains examples of hateful language some readers might find offensive.}
\end{abstract}

\section{Introduction}
Though once considered a problem driven primarily by reduced inhibitions in anonymous online spaces~\cite{rosner_verbal_2016, suler_online_2004}, offensive content has grown exponentially--to the point that many users no longer feel restricted by traditional conversational norms of tolerance and politeness, even when posting in their own names~\cite{rossini_beyond_2022}. The pervasiveness of toxic discourse on social media in particular has helped sow the seeds of discord and hatred that harm the health and well-being of its targets and pose significant threats to the fundamental rights of individuals and social groups on the margins~\cite{gelber2016evidencing}.
Concerned about such outcomes, over the last two decades scholars and practitioners from a variety of fields have scrutinized online incivility and hateful speech. Those working in natural language processing, for example, have developed techniques to detect different types of offensive discourse, ranging from incivility to hate speech, while social scientists have focused extensively on the larger substantive effects of these phenomena. 
%As is often the case, the two sides do not talk much. 

Most computational approaches for detecting online toxicity are based on classifiers that predict the presence of a \textit{single} main binary label~\cite{basile-etal-2019-semeval,stoll_detecting_2020, davidson_developing_2020, davidson_automated_2017}, with some notable exceptions~\cite[inter alia]{vidgen_introducing_2021,vidgen2021learning,mollas2022ethos,kennedy2022introducing}.\footnote{We refer to the works by \newcite{vidgen2020directions} and \newcite{poletto2021resources} for in-depth surveys.} However, while single-label binary approaches to harmful speech detection are conceptually tidy and tend to yield good predictive performance, they have major limitations. Most notably, such approaches \textit{are unable to distinguish discourse that threatens democratic norms, values, and rights from expressions that are merely rude or impolite.} Prior work detecting incivility, for instance, has combined relatively harmless expressions that break traditional norms of polite speech--for instance, profanities and swearing--with discourse that is potentially more harmful, such as personal insults, stereotyping, or hateful speech~\cite{stoll_detecting_2020,theocharis_bad_workman_2016,tang_dalzell_2019}. 
Binary approaches to toxic and offensive content detection oversimplify these complex concepts, and ultimately undermine researchers' and practitioners' ability to understand potential harms and evaluate what content should receive most focus and intervention, including for the purposes of content moderation.

To address these open issues, we show that our \textit{multi-label approach} rooted in insights drawn from social science is not only potentially more insightful, but also improves performance of detection models. In contrast to most previous work, we build upon a \textit{conceptual model that disentangles uncivil from intolerant online discourse}~\cite{rossini_beyond_2022}. The resulting labels can meaningfully distinguish discourse that is simply rude or offensive (\textit{incivility}) from expressions that threaten democratic norms and values, such as equality, diversity, and freedom (\textit{intolerance}).

We collect a dataset of more than 40,000 US- and UK-based tweets related to the topic of immigration, and annotate these tweets for four sub-types of \textit{incivility} (profanities, insults, outrage, character assassination) and two sub-types of \textit{intolerance} (discrimination, hostility). We refer to this dataset as \textit{Not Just Hate} (NJH). We then fine-tune large pre-trained language models and show that these labels can be predicted with consistently good performance. We compare our results to other benchmark hate speech datasets to produce more insights about the dataset we introduce. Models trained on our data match or outperform state-of-the-art performance on those datasets.

Our approach, annotation methodology, and dataset can help the future development of automated harmful online speech detection, and foster a more nuanced understanding of the distinctive types of discourse that constitute online toxicity and abuse. Data and additional details on the annotation are available on OSF.\footnote{\url{https://osf.io/gxvsj/?view_only=12197981e47a47239a6f80c62db84b14}} Details are also available on the GitHub repository.\footnote{\url{https://github.com/vinid/not-just-hate}}

\paragraph{Contributions}
We describe a novel perspective on harmful online speech detection. We describe in detal our annotation pipeline and we release a dataset, NJH, of just over 40,000 tweet ids annotated with four sub-types of incivility (profanities, insults, outrage, character assassination) and two sub-types of intolerance (discrimination, hostility). We show that our data set generalizes to various types of offensive language and reaches state-of-the-art performance.

\section{Data}

\paragraph{Collection}
We collected our dataset via the Twitter Enterprise API, downloading over 150 million tweets over the course of 2020-2021. We selected the keywords used to collect these tweets in a multi-stage process. Beginning with a list of 30 keywords and phrases commonly associated with immigration in the US and/or UK (e.g., immigration, immigrant, refugee, illegals), we drew a random sample of tweets containing those words from the public streaming API, produced a list of words commonly co-occurring with the seeded terms, and qualitatively analyzed that list to identify the appropriateness of each additional keyword or phrase. We performed the same process with a series of subreddits related to immigration, carefully curating the list of subreddits to represent both  pro- and anti-immigration sentiment, as well as a variety of immigration subtopics (e.g., subreddits dedicated to Asian or Muslim immigrants/immigration, refugees, etc.). Please see the Appendix for the final set of keywords used to collect tweets. 

\paragraph{Annotation}

Our annotation approach was based on quantitative content analysis \cite{krippendorff2018content}, a social scientific method used by communication scholars to interpret meaning in textual data at scale. The annotation guidelines were broadly inspired by those of \newcite{rossini_beyond_2022}, which we augmented and adapted with examples that were context- and country-specific to capture the nuances of immigration debates across the US and the UK. 

The annotators were ten undergraduate researchers from the University of Liverpool (UK) and Syracuse University (US). We trained the students on the annotation guidelines until they achieved a satisfactory inter-annotator reliability score for two consecutive weeks (Krippendorff's $\alpha$ of 0.68 or above) and Gwet's AC1 (of .6 or above). We used these to correct for expected issues in the data quality--Krippendorff's $\alpha$ penalizes data scarcity, which is a problem in some of our labels, while Gwet's AC1 corrects for the probability that the annotators agree by chance (this is more likely for difficult annotation tasks such as this one). We continuously monitored the quality of the annotation by measuring inter-annotator reliability on a monthly basis. 

How to aggregate annotation scores is a central and open problem in machine learning~\cite{Gordon2022JuryLI,davani-etal-2022-dealing}.
We divided the trained annotators into teams % of two or three 
to individually annotate tweets from their respective country.  We compared their individual annotations, and they met to discuss and adjudicate any disagreements--i.e., all annotators had to agree on a best label(s). We opted for this rigorous annotation process instead of a simple majority rule due to the complexity of the phenomena we investigate. While prior efforts on incivility and hate speech detection have relied on approaches such as 'majority rule' to determine labels based on crowdsourced annotations, leading to biases in interpretation that may undermine the quality of annotation, we leverage social scientific content analysis techniques to develop a rigorous annotation pipeline to ensure the quality of our dataset.

We randomly selected samples for annotation, stratifying by location. We restricted the data to the US and UK via geolocated tweets and user location, and use simple trigram tf-idf based classifier to distinguish between the UK, US, and Other (Macro F1=0.82, albeit we have updated our algorithm several times during the course of the research project) trained on 3 million tweets. Given the relative scarcity of intolerant discourse in particular, we used Jigsaw's Perspective API to over-sample potentially toxic tweets and increase the presence of some labels on approximately 60\% of the annotated sample.
% Figure~\ref{fig:taxonomy} shows the taxonomy of the labels used in this dataset;
Table~\ref{tab:numbers:dataset} shows the number of tweets tagged with each label in the dataset. %The most frequent label is \textsc{Discrimination}. Figure~\ref{tab:examples} shows some example of the tweets.

\begin{table}[] 
    \centering
    \begin{tabular}{l|c}  \toprule
        \textbf{Label} & \textbf{Number} \\ \midrule
        Outrage (O) & 6,743 \\
        Insults (I) &  5,040 \\ 
        Profanity (P) & 4,074 \\
        Char. Assassination (C) & 3,436 \\
        Discrimination (D) & 10,437 \\
        Hostility (H) & 2,699 \\
        No Label & 22,007 \\ \hline
        Total Labels:  & 57,139 \\
        \bottomrule
    \end{tabular}
    \caption{40,136 tweets. \textit{No Label} = tweets with no labels. \textit{Total Labels} includes tweets with multiple labels.}
    \label{tab:numbers:dataset}
\end{table}

% \begin{figure*}[ht!]
%   \centering  \includegraphics[width=1\textwidth]{img/taxonomy_incivility_intolerance.pdf}
%   \caption{This Figure describes the taxonomy of the annotations we have defined and used in this work. On Level 1, we have two main labels (incivility and intolerance). Each of these has several sub-classes in turn (Level 2).}
%   \label{fig:taxonomy}
% \end{figure*}

\begin{table}[ht!]
    \centering
    \small
    \begin{tabular}{p{0.6\columnwidth}|p{0.3\columnwidth}} \toprule
    \textbf{Example} & \textbf{Labels}\\
        \toprule
\small        @user and i don't trust dead liberal voters and illegals voting for more handouts. EC is where its at. Screw them lib frauds. &  \small\textsc{profanity, insults, character assassination, outrage, discrimination} \\ 
        \midrule
     \small   FUCK ICE! & \small\textsc{profanity, insults} \\
    %     \midrule
    % \small    French police clear migrant camp w more than 2000 people URL via @user Europe overflows w invaders, they must be sent home. Can always improve their own countries. & \small \textsc{outrage, hostility, discrimination} \\
        \bottomrule
    \end{tabular}
    \caption{Examples of tweets in our dataset with the respective labels. Text altered to preserve privacy.}
    \label{tab:examples}
\end{table}

\section{Experiments}

\subsection{Datasets}\label{sec:dataset}

\paragraph{HateEval (2,971 examples, 2 labels)} This dataset was introduced during the SemEval2019~\cite{basile-etal-2019-semeval} challenge, and it is particularly well-suited for our task because it has been built around the topics of immigration, which we cover, and women. Note that we use the samples and splits provided by the recently introduced TweetEval benchmark~\cite{barbieri2020tweeteval}.

\paragraph{HateCheck (421 examples, 2 labels)} This dataset offers a checklist to evaluate and stress test different hate speech detection models. Examples in HateCheck address linguistic features like spelling variation or negation (e.g., ``There is no hatred in my heart for you immigrants'' is a non hateful example). HateCheck is an excellent dataset to verify how a model behaves when encountering these features. However, examples are manually generated and cannot be considered \textit{real} examples of harmful speech. HateCheck covers different targets of harmful speech; however, we extract only the subset containing data related to immigration~\cite{rottger-etal-2021-hatecheck}.

\paragraph{Data Preparation}
We lighty pre-process the data: we replace user tags with an anonymous \textit{USER} and links with \textit{HTTPURL}.\footnote{Note that this is a common approach for Twitter data in large language models~\cite{bertweet}.} This is done to prevent the model from learning spurious patterns regarding the occurrence of specific users. We split our dataset into 3 sets: train (85\%, 34,115 examples), dev (7.5\%, 3011 examples), and test (7.5\%, 3011 examples).  

\begin{table*}[ht!]
    \centering
    \begin{tabular}{l|ccc|c} \toprule
        \textbf{Model} & \textbf{NJH} (Macro-F1) & \textbf{HateEval} (Macro-F1) & \textbf{HateCheck} (Accuracy)  & \textbf{Avg.}\\ 
        \midrule
        Roberta-base & 0.74 $\pm$ 0.00 &0.63 $\pm$ 0.01&0.54 $\pm$ 0.03  &  0.64\\
        Roberta-large & 0.76 $\pm$ 0.00 & \textbf{0.65} $\pm$ 0.02 & 0.69 $\pm$ 0.03 &  0.70\\
        BERTweet-base & 0.74 $\pm$ 0.01&0.64 $\pm$ 0.01&0.47 $\pm$ 0.06 &  0.62\\
        BERTweet-large & \textbf{0.77} $\pm$ 0.00&0.63 $\pm$ 0.02& \textbf{0.71} $\pm$ 0.02 &  0.70\\ 
        \midrule
        Best Other & -- & 0.52 $\pm$ 0.00 & \textbf{0.71} $\pm$ NA &  --\\ 
        \bottomrule
    \end{tabular}
    \caption{Comparison of various models trained on our data and tested on several data sets. On HateCheck we evaluate on Accuracy as in the original paper by~\newcite{rottger-etal-2021-hatecheck}.}
    \label{tab:results:dataset}
\end{table*}

\subsection{Models and Training}

We use RoBERTa~\cite{liu2019roberta} (base and large) and BERTweet~\cite{bertweet} (base and large). BERTweet is a RoBERTa model additionally pretrained on Twitter data. Each model is fine-tuned three times, we report averaged results.

While we annotate for six labels, during training we let the classifier also predict the \textit{supertypes} of the labels: \textit{incivility} for \textsc{profanity, insults, char. assassination, and outrage} and \textit{intolerance} for \textsc{discrimination and hostility}. The total number of labels to predict is thus eight.
% This is similar to approaches in Fine-Grained Entity Typing~\cite{onoe2020fine}. 

See the Appendix for the hyper-parameters used. We run a small parameter selection pipeline testing different learning rates \{5e-4, 5e-5, 5e-6\}. All models are trained for five epochs, but we select the model that performs best at validation time; validation is run every 200 steps. The learning rate of 5e-5 was the best performing, but we report all the results in the Appendix. We test all the trained models on the test portion of NJH and on the test set from HateEval and on HateCheck. We also report the best results on HateEval and HateCheck as described in the papers~\cite{barbieri2020tweeteval,rottger-etal-2021-hatecheck} (marked as \textit{Best Other} in Table \ref{tab:results:dataset}).

Both HateEval and HateCheck focus on the binary hate/not hate annotations. To adapt to the binary setting, since models trained on NJH are multi-label, we consider a tweet \textit{hateful} if the model predicts one or more of the following labels: \textit{hostility}, \textit{discrimination} and/or \textit{intolerance}. Note that because these datasets have been annotated with different definitions of \textit{hate}, results might not always be perfectly comparable.

\subsection{Results}
Table~\ref{tab:results:dataset} shows the results of our fine-tuned models on the different datasets. 

\begin{figure}[ht!]
  \centering  \includegraphics[width=1\columnwidth]{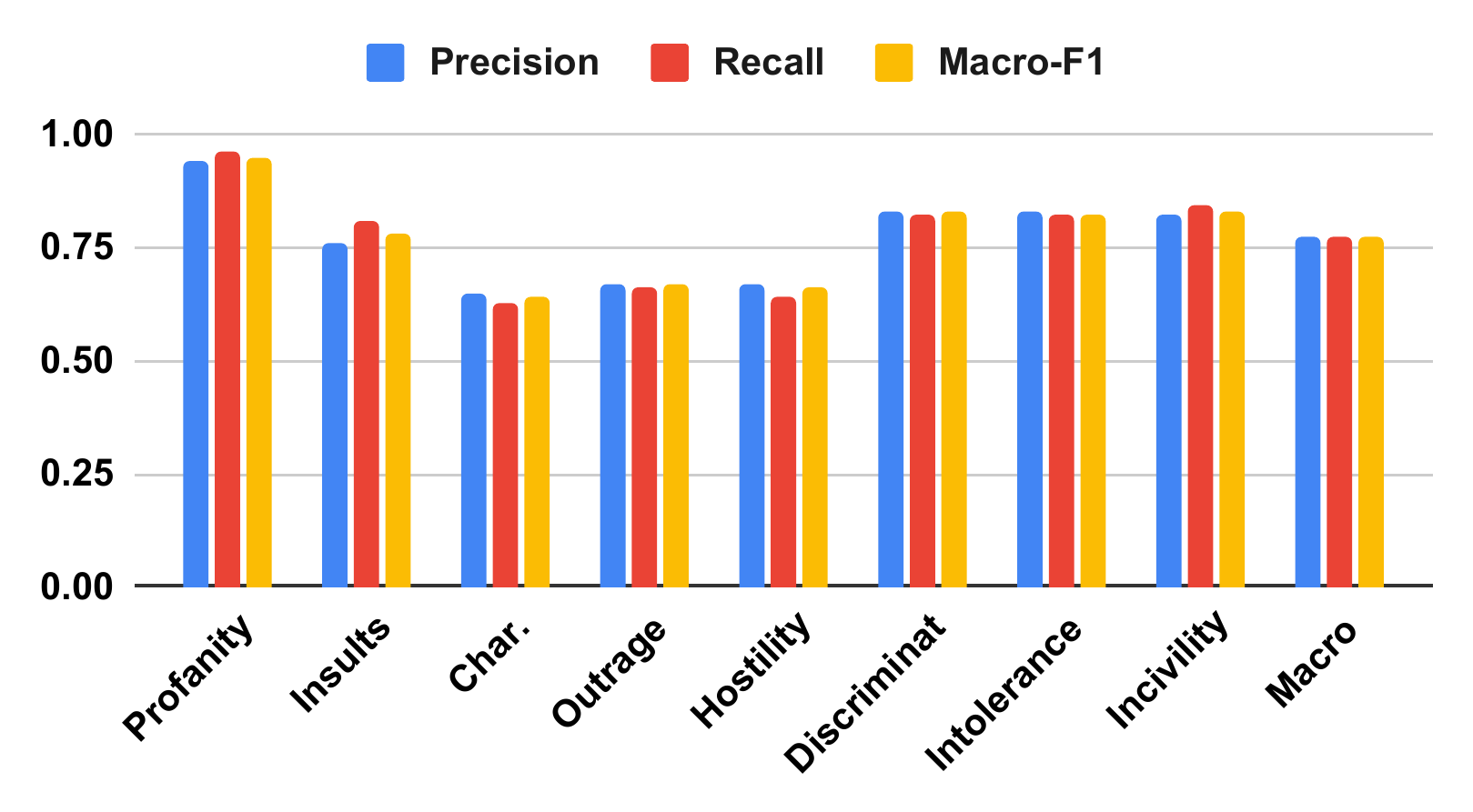}  \caption{F1 results per-label and Macro F1.}
  \label{fig:perlabel}
\end{figure}

\paragraph{NJH}
Performance for all models is consistently above 0.70 Macro F1. Figure~\ref{fig:perlabel} shows the results per label for the best model, BERTweet-large.

\paragraph{HateEval}
Models trained on our dataset achieve better Macro-F1 on HateEval than previous work~\cite{barbieri2020tweeteval}, reaching results comparable to those in the challenge~\cite{barbieri2020tweeteval}.
Best Other is described by~\cite{barbieri2020tweeteval}, a RoBERTa model, fine-tuned on the HateEval training data~\cite{basile-etal-2019-semeval}.

\paragraph{HateCheck}

Performance on the immigration subset suggests that the base models do not learn as well as the large ones. However, BERTweet-large reaches a comparable performance to the best model.
Best Other is the described by~\cite{rottger-etal-2021-hatecheck}, a BERT model fine-tuned on the Twitter dataset by~\newcite{davidson_automated_2017}.

\subsubsection{Comparison with other Models}
\begin{figure}[ht!]
  \centering  \includegraphics[width=1\columnwidth]{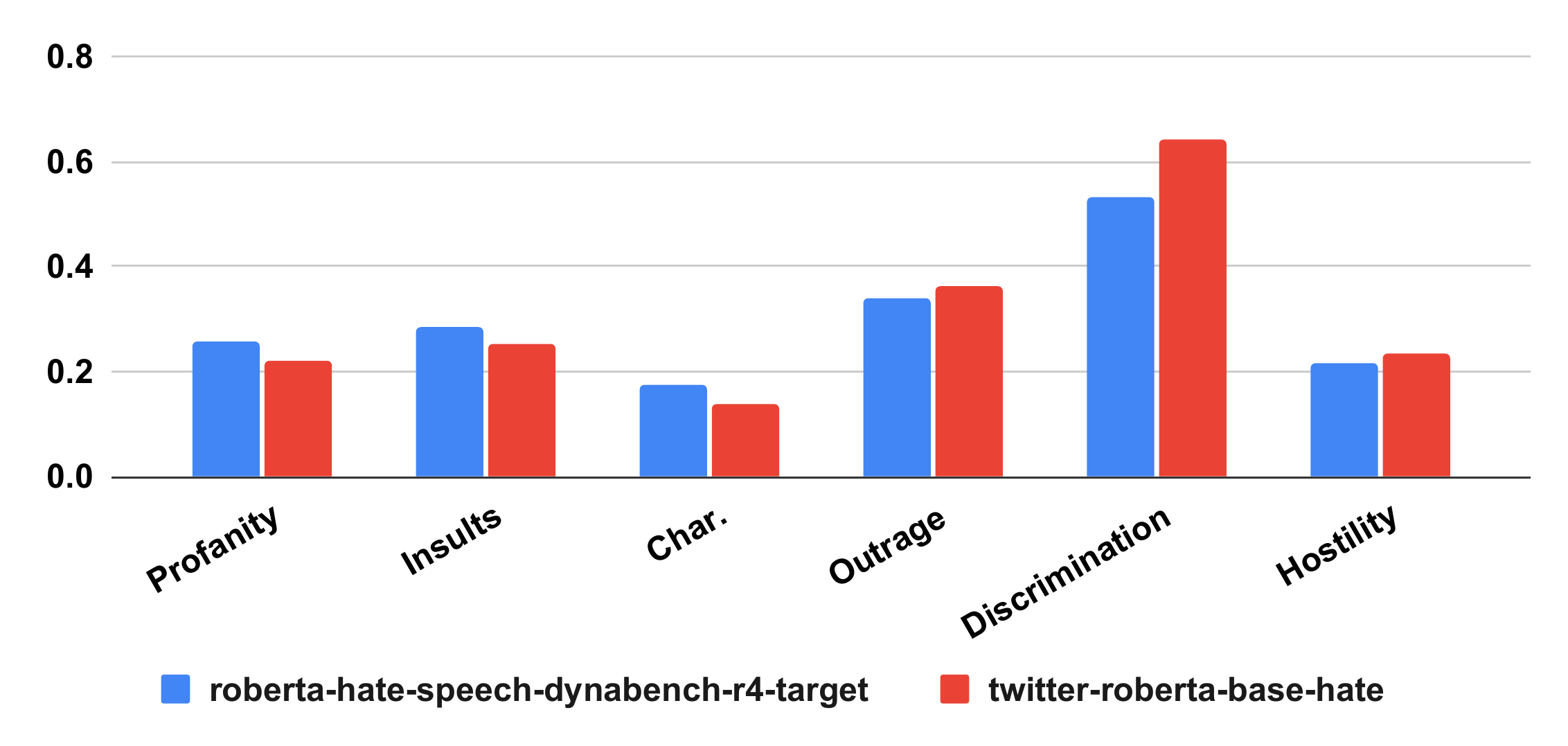}
  \caption{F1 of existing models on our dataset.}
  \label{fig:njhvsmodels}
\end{figure}

We compare the performance of other pre-trained models on NJH. This serves as a proof of concept that popular approaches do not capture the entire spectrum of incivility and intolerance proposed by our data. We use two models: one trained on data collected from different rounds of human-machine interaction to train better models~\cite{vidgen2021learning}.\footnote{\url{https://huggingface.co/facebook/roberta-hate-speech-dynabench-r4-target}} and one\footnote{\url{https://huggingface.co/cardiffnlp/twitter-roberta-base-offensive}} that has been trained on HateEval data~\cite{barbieri2020tweeteval,basile-etal-2019-semeval}.  We then compute the F1 score between the predictions of the models and each of our labels. Figure~\ref{fig:njhvsmodels} shows the results. Both models seem to be able to effectively capture \textsc{Discrimination}; however they do not capture stronger harmful speech such as \textsc{Hostility}. In general, the models do not seem to capture \textbf{Incivility} (\textsc{Profanity, Insults, Outrage}, or \textsc{Character assassination}). This latter result can be expected since the models have been trained to predict just hateful content. Overall, our findings suggest that there is a need for models that can effectively distinguish different aspects of offensive and potentially harmful discourse.

% We evaluate the performance of models trained our proposed dataset (NJH) as well as on HateEval~\cite{basile-etal-2019-semeval} and on HateCheck~\cite{rottger-etal-2021-hatecheck}. Both of these datasets contain data on immigration, in line with our collected data.

%\section{Guidelines for Future Work}

\section{Conclusions}

We suggest that a more fine-grained approach improves offensive and potentially harmful speech detection online, and, crucially, can allow for a better understanding of the spectrum of online toxicity. Our approach can successfully disentangle incivility from likely more harmful cases of intolerant content, allowing scholars and practitioners to better understand and detect discourse that undermines broader democratic norms, values, and rights~\cite{rossini_beyond_2022}. We introduce a dataset of just over 40,000 tweets, annotated with six labels. Models trained on our dataset can predict labels with good confidence and perform well on other benchmark datasets.

\section*{Acknowledgments}
DH is a member of the Bocconi Institute for Data Science and Analytics (BIDSA). This work was partially conducted while FB was a member of BIDSA. 

We acknowledge financial support for this research in the form of a gift from Twitter, Inc. 

\section*{Ethical Considerations}
We anonymized Twitter handles as part of the data pre-processing, and any tweet text provided as an example here (i.e., in Table~\ref{tab:examples}) has been edited to further preserve anonymity. The NJH dataset is shared in dehydrated format, i.e., as tweet IDs only, in full compliance with Twitter's Developer Policy.\footnote{\url{https://developer.twitter.com/en/developer-terms/policy}} 
We are aware that our dataset, if reconstructed, contains potentially harmful content. Though we use this content to help better examine, understand, and help mitigate the harms of online hate, we recognize that these tweets could be used for darker purposes. As any tweet successfully rehydrated from our list of tweet IDs remains in the public domain, we have assessed that the benefits of sharing this dataset outweigh the risks.

\section*{Limitations}

\textsc{Hostility} is an aggregated label that encompasses the originally annotated labels of \textsc{Hateful Speech}, \textsc{Dehumanization}, \textsc{Serious Threat-Personal Abuse-Harassment}, and \textsc{Democratic Threat}. These labels did not yield enough annotated tweets to remain a part of our multi-label classifier in their own right. Although \textsc{Hostility} is a suitable label that groups the original labels on the basis of hostile intent and/or effect, as well as the nature of their targets, we cannot claim that annotators would have interpreted tweets in the same way if they had annotated for \textsc{Hostility} rather than following codebook guidance for each individual label. For full transparency, we are releasing all original, ungrouped, annotations for this dataset. 

Although we are also releasing the unaggregated annotations alongside the  aggregated annotations, it must be noted that the nature of our adjudication process means that our aggregated labels cannot be directly reproduced from the unaggregated ones. This is because we opted for a significantly more rigorous approach that involved annotators meeting to discuss and resolve every single annotation disagreement under expert supervision. At times, these discussions might have led to the decision to annotate for labels that previously no individual annotator had identified. Although more rigorous and fair - through ensuring every annotator's views are heard - this process has the downside of being less transparent retrospectively, as the discussion and decision-making that took place in these adjudication meetings cannot be easily documented and the final aggregated annotations ultimately only present the outcome of the process and not the process itself. 

An additional limitation to replicability stems from the decay of tweets over time, wherein deleted tweets and/or tweets from suspended, deleted, or newly private accounts cannot be rehydrated based on their tweet IDs. This is a common event in all Twitter datasets~\cite{tromble2017lost}, and is particularly prevalent in hate speech datasets, where users are often suspended and individual tweets removed from the platform. We currently estimate approximately 25\%  of tweets in this dataset are no longer accessible for rehydration.

% Entries for the entire Anthology, followed by custom entries
\bibliography{anthology,custom}
\bibliographystyle{acl_natbib}

\appendix

\section{Dataset Details}

\subsection{Data Statement}
The data we share is is composed by tweet ids and does not directly contain personal information of the individual; however upon reconstruction, it shows tweet and author if it is still publicly available. The reconstructed data contains harmful messages. Annotators were all English native speakers.

\subsection{Twitter Keyword Used}

In the following we include the list of keywords used to extract the tweets.

"the wall" OR "fuck ice" OR undocumented OR illegals OR "an illegal" OR "muslim ban" OR "travel ban" OR refugee OR asylum OR \#wherearethechildren OR "child cage"$\sim$3 OR "children cage"$\sim$3 OR \#wall OR daca OR \#dreamer OR "sanctuary city" OR "sanctuary cities" OR "baby cage"$\sim$3 OR "babies cage"$\sim$3 OR "abolish ice"$\sim$3 OR "ice raid"$\sim$3 OR \#abolishice OR \#muslimban OR \#nobannohate OR \#refugeeswelcome OR \#refugeeswelcomehere OR ms-13 OR "build the wall" OR \#buildthewall OR “ms- 13” OR "ms 13" OR deport OR citizenship OR birthright OR "illegal alien"$\sim$3 OR ms13 OR \#secureourborders OR \#familiesbelongtogether OR \#closethecamps OR \#defenddaca OR \#nocamps OR \#noban OR \#savedaca OR \#immigrationreform OR \#uslatino OR \#openborders OR "open border"$\sim$3 OR "kid cage"$\sim$3 OR "kids cage"$\sim$3 OR USCIS OR \#proimmigration OR "farm worker" OR "farm workers" OR farmworker OR \#farmworkerjustice OR \#immigrationpolicy OR migrant OR amnesty OR \#noamnesty OR \#imalreadyhome OR \#proopenborders OR \#immigrantnation OR \#nohumanisillegal OR \#welcomeimmigrants OR "no human is illegal" OR \#MSW52170 OR \#immigrantsmatter OR \#immigrantrights OR "learn to speak English" OR "steal jobs" OR "job stealing"$\sim$3 OR "mexican border" OR "mexico pay"$\sim$3 OR visa OR "chain migration" OR "dream act" OR "merit based" OR citizen OR foreigner OR "foreign national" OR "trump wall"$\sim$3 OR "mexico policy"$\sim$3 OR "foreign worker"$\sim$3 OR "human trafficking"$\sim$3  OR xenophobe OR xenophobia OR schengen OR "british national" OR \#BNO OR "free movement" 

% \subsection{Examples}

% \begin{figure*}[ht!]
%   \centering  \includegraphics[width=0.8\textwidth]{img/incivility_intolerance_examples.pdf}
%   \caption{Example of tweets in our dataset with the respective labels.}
%   \label{fig:examples}
% \end{figure*}

% \section{Model Training}
% Other results

\section{Model Training}

Results on validation set are available in Figure~\ref{tab:eval}

\begin{table}[ht]
    \centering
    \begin{tabular}{l|cc}\toprule
    Model & 5e-5 & 5e-6 \\ \midrule
         Roberta-base    & 0.74 $\pm$ 0.01 & 0.67 $\pm$ 0.01\\
         Roberta-large & 0.76 $\pm$ 0.00 & 0.74 $\pm$ 0.00 \\
         BERTweet-base  & 0.73 $\pm$ 0.01& 0.54 $\pm$ 0.02 \\
         BERTweet-large  & 0.77 $\pm$ 0.00 & 0.76 $\pm$ 0.00\\  \bottomrule
    \end{tabular}
    \caption{Results on the NHJ validation set. *models with a learning rate of 5e-4 obtained very low performance and were not working on the data.}
    \label{tab:eval}
\end{table}

Figure~\ref{tab:training:details} shows the parameters used to train the models  (excluding learning rate that is a parameter we found with grid search).

\begin{table}[ht]
    \centering
    \begin{tabular}{l|c} \toprule
        \textbf{Param} &  \textbf{Value} \\ \midrule
        Batch Size & 64 \\
        Learning Epochs* & 5 \\
        Optimizer & AdamW \\ 
        Betas & 0.9 and 0.999 \\
        Max Length & 80 \\ \bottomrule
    \end{tabular}
    \caption{The main parameters we used to run the models. *While epochs are 5, we remark that we are running a step-wise evaluation. Batch size is achieved thanks to the use of gradient accumulation (8 steps)}
    \label{tab:training:details}
\end{table}

\end{document}